\documentclass[conference]{IEEEtran}
\IEEEoverridecommandlockouts

\usepackage{amsmath,amsfonts,bm}

\def\eqref#1{equation~\ref{#1}}

\def\1{\bm{1}}

\DeclareMathAlphabet{\mathsfit}{\encodingdefault}{\sfdefault}{m}{sl}
\SetMathAlphabet{\mathsfit}{bold}{\encodingdefault}{\sfdefault}{bx}{n}

\usepackage{amsmath,amssymb,amsfonts}
\usepackage{algorithmic}
\usepackage{graphicx}
\usepackage{textcomp}
\usepackage{xcolor}

\usepackage[bookmarks=false]{hyperref}
\usepackage{url}

\usepackage[linesnumbered,ruled,vlined]{algorithm2e}
\usepackage{xcolor}

\SetCommentSty{mycommfont}

\SetKwInput{KwInput}{Input}                
\SetKwInput{KwOutput}{Output}              

\usepackage{mathtools}
\usepackage{verbatim}
\usepackage{subcaption}

\usepackage{cite}

\begin{document}

\title{Bias-Free FedGAN: A Federated Approach to Generate Bias-Free Datasets\\}

\author{

\IEEEauthorblockN{Vaikkunth Mugunthan*\thanks{ * Equal Contribution}}
\IEEEauthorblockA{\textit{MIT} \\
Cambridge, USA \\
vaik@mit.edu}

\and

\IEEEauthorblockN{Vignesh Gokul*}
\IEEEauthorblockA{\textit{UCSD} \\
San Diego, USA \\
vgokul@ucsd.edu}

\and

\IEEEauthorblockN{Lalana Kagal}
\IEEEauthorblockA{\textit{MIT} \\
Cambridge, USA \\
lkagal@mit.edu}

\and

\IEEEauthorblockN{Shlomo Dubnov}
\IEEEauthorblockA{\textit{UCSD} \\
San Diego, USA \\
sdubnov@ucsd.edu}

}

\maketitle

\begin{abstract}
Federated Generative Adversarial Network (FedGAN) is a communication-efficient approach to train a GAN across distributed clients without clients having to share their sensitive training data. In this paper, we experimentally show that  FedGAN generates biased data points under non-independent-and-identically-distributed (non-iid) settings. Also, we propose Bias-Free FedGAN, an approach to generate bias-free synthetic datasets using FedGAN. Our approach generates metadata at the aggregator using the models received from clients and retrains the federated model to achieve bias-free results for image synthesis. Bias-Free FedGAN has the same communication cost as that of FedGAN. Experimental results on image datasets (MNIST and FashionMNIST) validate our claims.
\end{abstract}


\section{Introduction}

Generative adversarial networks (GANs) \cite{goodfellow2014generative}, a  class of generative models \cite{rezende2014stochastic,li2015generative}, are used to generate synthetic datasets that are similar to the datasets in which they were trained on. Application of GANs include generating video from images, style transfer \cite{azadi2018multi}, improving resolution of pictures \cite{ledig2017photo}, creating deepfake videos \cite{korshunov2018deepfakes}, etc. High-quality GANs are trained on large datasets. However, in most cases, data is distributed across different sources. Sharing sensitive data is not an option due to regulations such as GDPR \cite{EUdataregulations2018}, CCPA \cite{de2018guide}, and HIPPA \cite{hipaa}. Hence, to train the entire population, a distributed GAN approach is required. \cite{hardy2019md} proposed a system that has a single generator and distributed discriminators. The discriminators exchange their model parameters to avoid overfitting. A distributed training paradigm using GANs that automatically learns feature-control variables was proposed by \cite{mugunthan2020dpd}. However, these approaches work only for iid data sources. The system proposed by \cite{yonetani2019decentralized} works for non-iid data sources. In their algorithm, individual discriminators are trained and they update a centralized generator. However, their work faced numerous communication challenges. 
To solve this problem, FedGAN (Federated GAN) was proposed by \cite{rasouli2020fedgan}. In FedGAN, multiple clients collaboratively train a model (generator and discriminator pair) without sharing their raw data; they share their local model weights with a trusted aggregator during each round of training; the aggregator updates the global model (generator and discriminator pair) using these weights. This process repeats for a pre-defined number of rounds or until convergence is achieved. 

Despite proving convergence and producing high-quality results, FedGAN generates biased data in multiple scenarios. The authors of FedGAN \cite{rasouli2020fedgan} do not address this major issue.  Though there is a myriad of solutions for addressing fairness and bias in GANs under the local setting
\cite{sattigeri2018fairness, xu2018fairgan, tan2020improving, hwang2020fairfacegan}, no solutions have been proposed for solving this problem in the federated setting. 

One of the main features of Federated Learning is the heterogeneity of the clients' data. This also causes biases in the results, as it is not possible to manually evaluate the bias or access the images of the clients. Some approaches
\cite{yao2019federated, huang2020fairness, abay2020mitigating} have been proposed to mitigate biases in federated learning. \cite{abay2020mitigating} use local and global reweighing and propose a fairness-aware regularization term in the training objective. \cite{huang2020fairness} propose a double momentum gradient method and a weighting strategy based on the frequency of participation in the training process. However, these approaches do not provide a solution for federated/distributed GANs. 

We consider an example scenario to emphasize the importance of this problem. Let us consider the task of face synthesis using GANs in a federated setting. The goal of the generative model is to synthesize realistic faces that mimic the training dataset. In a federated setting, each client is unaware of the other clients' data. For example, let us consider a scenario in which a particular client has a training dataset that has facial images of black people. Let the other clients have datasets belonging to another race. We demonstrate that federated models trained under such settings can propagate bias and ignore the minority class data points while synthesizing output images. This is a serious issue as these models can perform poorly and unethically for minority classes.

In this paper, we propose Bias-Free FedGAN, a training paradigm that eliminates biases in federated models. In our method, the aggregator generates a new dataset (metadata) from each of the incoming client models and retrains the federated model. We show experimental results that the combination of averaging weights of client models and retraining on metadata helps achieve bias-free results in image synthesis.

In summary, our contributions are as follows:
\begin{itemize}
    \item We demonstrate that FedGAN produces biased results under non-iid scenarios.
    \item We propose Bias-Free FedGAN, a framework to train FedGAN in an unbiased manner to produce bias-free outputs.
    \item We validate our claims by running experiments on the MNIST \cite{lecun2010mnist} and FashionMNIST \cite{xiao2017fashion} datasets. 
\end{itemize}

\begin{figure*}
\centering
  \includegraphics[width=\textwidth]{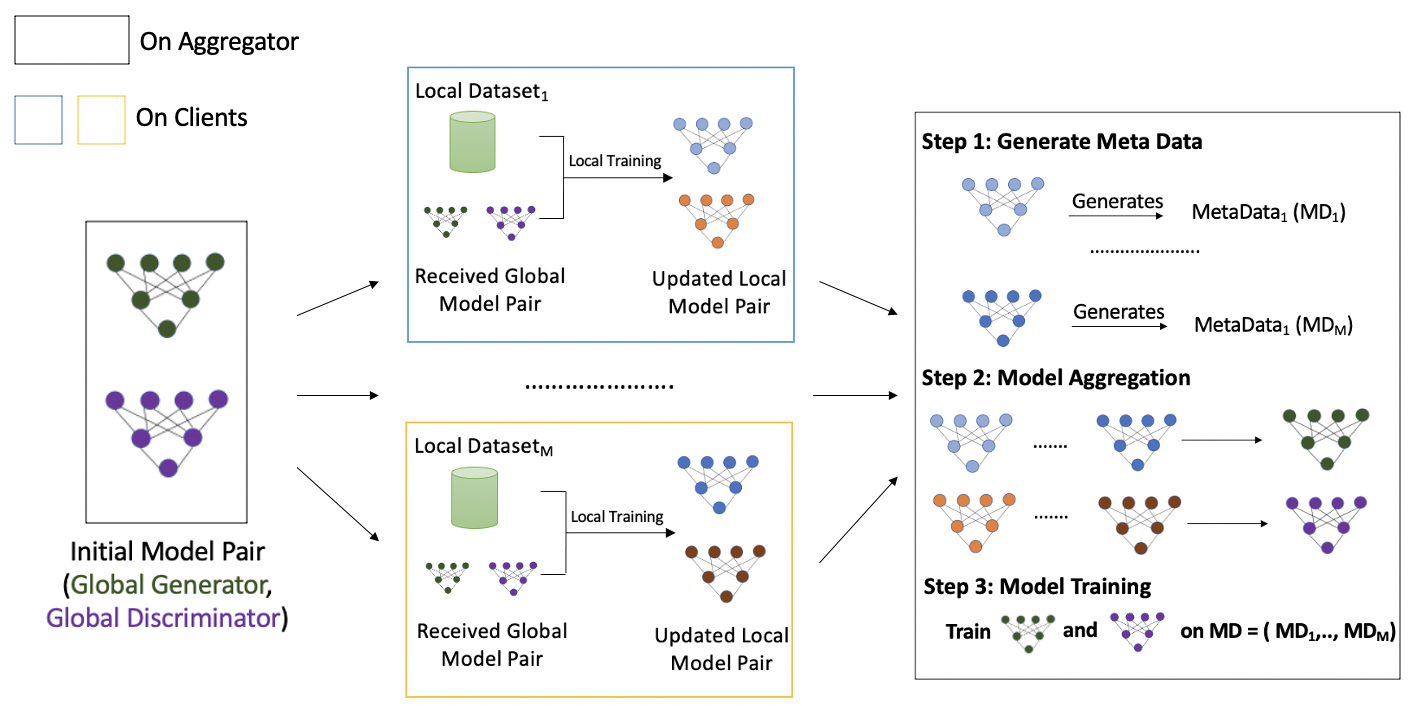}
      \caption{ A Representative Round in Bias-Free FedGAN}
    \label{fig:workflow}
\end{figure*}

\subsection{Paper Organization}
Section II provides a detailed explanation of GAN and FedGAN.  In Section III, we propose our Bias-Free FedGAN algorithm and explain why our approach works. Experiments and evaluations under different scenarios are described in Section IV. We conclude with a summary and a discussion of our future work in Section V. 

\section{Background}

\subsection{Generative Adversarial Networks}
Generative Adversarial Networks employ two networks that train simultaneously together. A generator network $G$ upsamples a latent space $p(z)$ to an image, while a discriminator $D$ tries to predict if a given image is from the generator or the real data distribution $p(x)$. This leads to a minmax game as follows:

\begin{align*}
 \min_{G} \max_{D} V(D,G) = E_{x \sim p_{(x)}}[log(D(x))] + \\ E_{z \sim p(z)}[log(1 - D(G(x))]
\end{align*}
The generator trains to fool the discriminator to make false predictions, while the discriminator tries to make strong predictions about real and fake data. At the beginning of training, the generator produces noisy images and the discriminator easily classifies between fake and real images. As training progresses, the generator synthesizes images that are closely similar to the real dataset and hence fools the discriminator.

\subsection{FedGAN}
The FedGAN framework was proposed by \cite{rasouli2020fedgan}. FedGAN is used to train a GAN across non-independent-and-identically-distributed data sources. Their system uses an aggregator for averaging and broadcasting the model parameters of the generator and discriminator. Algorithm \ref{ref:fedgan-algo} in the Appendix provides the pseudo-code for FedGAN. In addition, the authors also prove that FedGAN has a similar performance to the general distributed GAN while reducing communication complexity.

%
%




\section{Bias-Free FedGAN}

In this section, we propose out Bias-Free FedGAN algorithm.
We consider $M$ clients {1, 2, ..., M} and $n$ denote the index time. Each client $i$ has a local dataset $D_i$, generator $\boldsymbol \alpha_n^i$, and discriminator $\boldsymbol \beta_n^i$. $\eta_1$ and $\eta_2$ denote the learning rate for the generator and discriminator respectively.

We present the Bias-Free FedGAN algorithm in Algorithm \ref{ref:algos}. Each client runs the ADAM optimizer to train their local generators and discriminators on their datasets. As shown in Figure \ref{fig:workflow}, clients send their generator and discriminator model parameters to a central aggregator. The aggregator generates combined metadata ($MD_n$) using the generators received from the clients. For example, $MD_n$ can be collected by sampling 10000 images from each client's generator. The aggregator averages $\boldsymbol \alpha_n^i$ and $\boldsymbol \beta_n^i$ across $i$. Then, the aggregator trains the averaged model on $\mathbf{MD}$. The aggregator sends the final generator $\boldsymbol \alpha_n^g$ and discriminator $\boldsymbol \beta_n^g$ parameters to the clients. This process repeats for $N$ rounds.

\begin{algorithm}[h]
\DontPrintSemicolon

  \KwInput{Number of training rounds $N$. Initialize local generator and discriminator for each client $i$: $\boldsymbol \alpha_0^i=\boldsymbol \alpha'$ and $\boldsymbol \beta_0^i= \boldsymbol \beta'$, $\forall i \in {1, 2, ..., M}$. Learning rates of discriminator and generator, $\eta_1$ and $\eta_2$. Noise seed, $z$. }

  \For{n = 1 to N}
  {
    Each client $i$ computes local gradient $\bm{g}^i(\bm{\alpha}^i_n,\bm{\beta}^i_n)$ from $D_i$ and $\bm{h}^i(\alpha^i_n,\beta^i_n)$ from $D_i$ and synthetic data generated by the local generator.

    Each client $i$ updates its local model in parallel to other clients via 
    \begin{equation}
        \bm{\alpha}^i_n = \bm{\alpha}^i_{n-1} + \eta_1 \bm{g}^i(\bm{\alpha}^i_{n-1},\bm{\beta}^i_{n-1})
    \end{equation}
    \begin{equation}
        \bm{\beta}^i_n = \bm{\beta}^i_{n-1} + \eta_2 \bm{h}^i(\bm{\alpha}^i_{n-1},\bm{\beta}^i_{n-1})
    \end{equation}
    
    Each client $i$ sends model parameters $\bm{\alpha}^i_n$ and $\bm{\beta}^i_n$ to aggregator
     
     Aggregator generates metadata $MD_n$ by
     \begin{equation}
         MD_n = [MD^0_n, MD^1_n, MD^2_n, \dots ,MD^M_n]
     \end{equation}
     \begin{equation}
         MD^i_n = \bm{\alpha}^i_n(z)
     \end{equation}
     
     Aggregator computes global generator $\bm{\alpha}^g_n$ and global discriminator $\bm{\beta}^g_n$ by averaging
     \begin{equation}
        \bm{\alpha}^g_n = \sum_{j=1}^{M} \bm{\alpha}_n^j ; \bm{\beta}^g_n = \sum_{j=1}^{M} \bm{\beta}_n^j
    \end{equation}

    Aggregator trains $\bm{\alpha}^g_n$ and $\bm{\beta}^g_n$ on metadata $MD_n$

     Aggregator sends $\bm{\alpha}^g_n$ and $\bm{\beta}^g_n$ to clients and clients update 
     \begin{equation}
         \bm{\alpha}^i_n = \bm{\alpha}^g_n ; \bm{\beta}^i_n = \bm{\beta}^g_n , \forall i \in {1, 2, ..., M}
     \end{equation}

  }
  
\caption{Bias-Free FedGAN}
\label{ref:algos}

\end{algorithm}

\subsection{Why our approach works ?}

\begin{figure}

\centering
  \includegraphics[scale = 0.63]{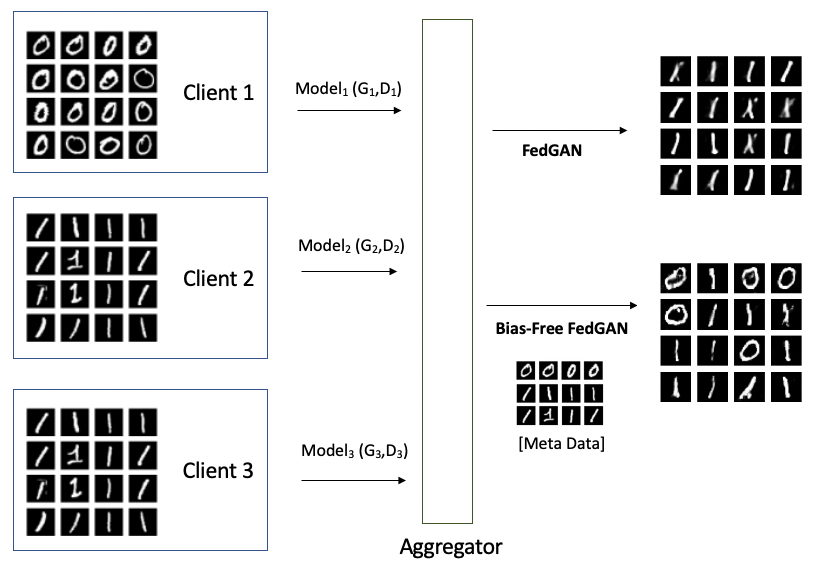}
      \caption{Sample Outcomes for FedGAN and Bias-Free FedGAN}
    \label{sample}
\end{figure}

We leverage the fact that each client's generator consists of an approximation of the real data distribution. We assume that each client has unbiased data. Since the aggregator has access to all the clients' trained generators, it also has access to an approximation of all the clients' data and can create metadata. The metadata consists of an equal number of images from all the generators. We retrain the federated model on a collection of all the generators' outputs to obtain a bias-free model. This process continues for each federated round. 

For example, consider the scenario where client 0 has class 0 and clients 1 and 2 have class 1 (as shown in Figure \ref{sample}). The outputs of the FedGAN model would be biased towards class 1. In Bias-Free FedGAN, there is an additional training step in the aggregator. The aggregator generates the metadata which would be a collection of images both belonging to class 1 and 0 in this case. The federated model is fine-tuned on the metadata before sending the model back to the clients. As the training progresses, the quality of the metadata improves and the Bias-Free FedGAN framework produces unbiased results without compromising the quality of images.

\section{Experiments and Analysis}

\begin{figure*}
     \centering
    \begin{subfigure}[t]{0.31\textwidth}
    
        \raisebox{-\height}{\includegraphics[width=\textwidth]{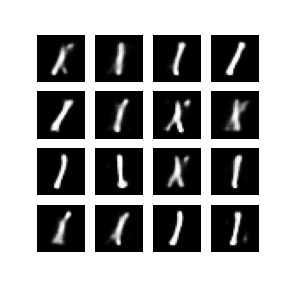}}
        \caption{}
        \label{3a}
    \end{subfigure}
    \hfill
    \begin{subfigure}[t]{0.31\textwidth}
        \raisebox{-\height}{\includegraphics[width=\textwidth]{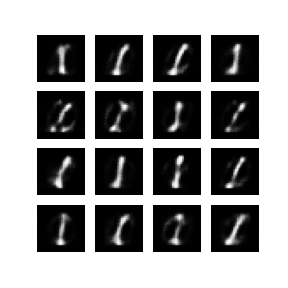}}
        \caption{}
        \label{3b}
    \end{subfigure}
    \hfill
    \begin{subfigure}[t]{0.31\textwidth}
        \raisebox{-\height}{\includegraphics[width=\textwidth]{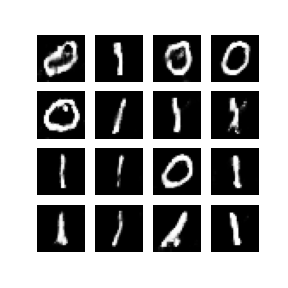}}
        
        \caption{}
        \label{3c}
    \end{subfigure}
    \begin{subfigure}[t]{0.31\textwidth}
        \raisebox{-\height}{\includegraphics[width=\textwidth]{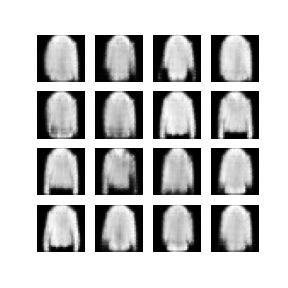}}
    \caption{} 
    \label{3d}
    \end{subfigure}
     \hfill
    \begin{subfigure}[t]{0.31\textwidth}
        \raisebox{-\height}{\includegraphics[width=\textwidth]{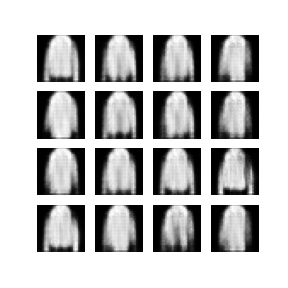}}
    \caption{} 
    \label{3e}
    \end{subfigure}
     \hfill
    \begin{subfigure}[t]{0.31\textwidth}
        \raisebox{-\height}{\includegraphics[width=\textwidth]{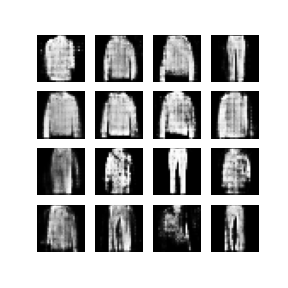}}
        
    \caption{} 
    \label{3f}
    \end{subfigure}
    \caption{(a-c) trained on MNIST; (d-f) trained on FashionMNIST; FedGAN (b,f) with client 1 trained on 10000 data points from the minority class and clients 2-5 trained on 10000 data points from the majority class; FedGAN (c,g) with client 1 trained on 10000 data points from the minority class and clients 2-5 trained on 2500 data points from the majority class; Bias-Free FedGAN (d,h) with client 1 trained on 10000 data points from the minority class and clients 2-5 trained on 2500 data points from the majority class.}
\end{figure*}

\begin{figure*}
     \centering
    \begin{subfigure}[t]{0.31\textwidth}
        \raisebox{-\height}{\includegraphics[width=\textwidth]{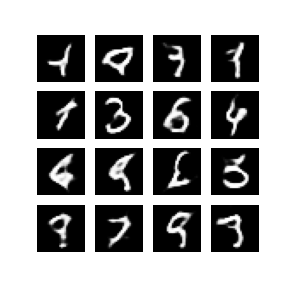}}
        \caption{}
        \label{4a}
    \end{subfigure}
    \hfill
    \begin{subfigure}[t]{0.31\textwidth}
    
        \raisebox{-\height}{\includegraphics[width=\textwidth]{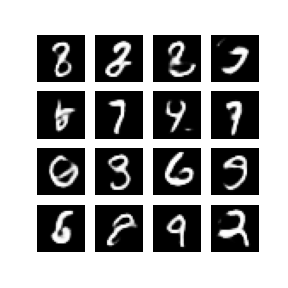}}
        \caption{}
        \label{4b}
    \end{subfigure}
    \hfill
    \begin{subfigure}[t]{0.31\textwidth}
        \raisebox{-\height}{\includegraphics[width=\textwidth]{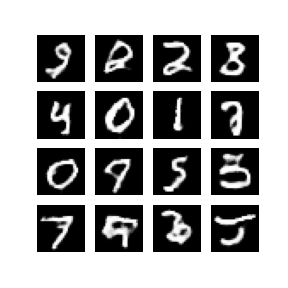}}
        \caption{}
        \label{4c}
    \end{subfigure}

\begin{subfigure}[t]{0.31\textwidth}
        \raisebox{-\height}{\includegraphics[width=\textwidth]{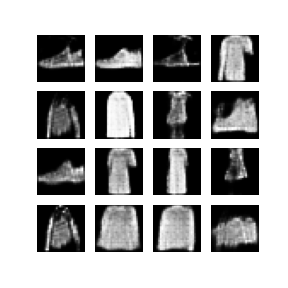}}
    \caption{}
    \label{4d}
    \end{subfigure}
    \hfill
    \begin{subfigure}[t]{0.31\textwidth}
        \raisebox{-\height}{\includegraphics[width=\textwidth]{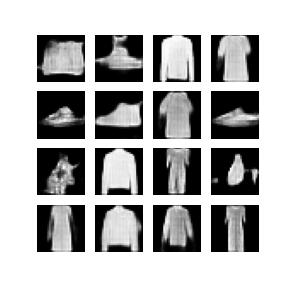}}
    \caption{} 
    \label{4e}
    \end{subfigure}
     \hfill
    \begin{subfigure}[t]{0.31\textwidth}
        \raisebox{-\height}{\includegraphics[width=\textwidth]{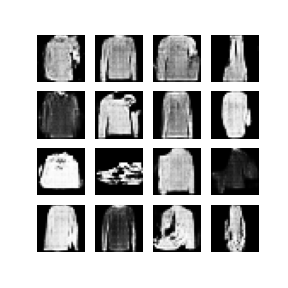}}
    \caption{} 
    \label{4f}
    \end{subfigure}
    \caption{(a-c) trained on MNIST; (d-f) trained on FashionMNIST; FedGAN trained in an iid setting (a,d);  FedGAN (b,e) with client 1 trained on data points from the minority classes 0,1 and clients 2-5 trained on data points from the majority class 2-9; Bias-Free FedGAN (c,f) with client 1 trained on data points from the minority classes 0,1 and clients 2-5 trained on data points from the majority classes 2-9.}
\end{figure*}

In this section, we show how FedGAN generates biased results and how Bias-Free FedGAN solves this problem. We demonstrate this by running experiments on the MNIST and FashionMNIST datasets. We use the same GAN architecture in all our experiments. The generator maps the latent space to the image space via a dense layer and three fractionally-strided convolutional layers. The discriminator consists of three convolutional layers and a dense layer with a sigmoid unit to discriminate between real and fake images. The number of epochs is set to 100 and the learning rate for the generator and discriminator is set to 0.0001. For the federated training, we set the number of federated rounds to 3.


Firstly, we show that FedGAN does not induce bias under the iid setting. We construct a setup where each client has an even and a balanced split of the training data. As shown in Figures \ref{4a} and \ref{4d}, we see that FedGAN generates all the classes in an iid setting.

Secondly, we analyze the federated setting in which a single client has the minority class and other clients have the majority class. Here, we consider a federated setting with 5 clients. Client 1 has images of the minority class (0 for MNIST and 'trouser' for FashionMNIST) and the other clients have images of the majority class (1 for MNIST and 'coat' for FashionMNIST). Specifically, client 1 contains 10000 images of the minority class and clients 2-5 contain 10000 images of the majority class. Figures \ref{3a} and \ref{3d} show that the output of FedGAN consists of images only from the majority class. That is, the aggregated model is heavily biased towards the majority class.

We also conduct experiments in which the total number of data points across both classes are equal, i.e, client 1 has 10000 images of the minority class and clients 2-5 have 2500 images of the majority class. Even in this scenario (Figures \ref{3b} and \ref{3e}), we observe that the generated images of FedGAN are heavily biased towards the majority class.

Additionally, we consider a scenario, where multiple minority classes exist. A single client has datapoints from classes 0 and 1, while other clients have images belonging to classes 2-9. Even in this scenario, we see that FedGAN (Figures \ref{4b} and \ref{4e}) omits classes 0 and 1 and is heavily biased towards the majority classes.

From prior experiments, we show that FedGAN can be biased towards the majority class under non-iid settings. Referring back to the face generation task, there is a high possibility that a federated model would generate white faces while ignoring minority races.

Finally, we show that Bias-Free FedGAN produces  unbiased results in the federated setting.
In this experiment, we set the number of clients to 5 and consider a non-iid setting across clients. Client 1 has 10000 images of the minority class, while clients 2-5 have 2500 images of the majority class. After each client shares the model parameters with the aggregator, the latter generates 10000 images from each client's generator to construct metadata of 50000 images. The aggregator trains an averaged model on the metadata for 100 epochs and sends the global model back to each client. 

Despite the class imbalance across clients, Bias-Free FedGAN produces results that are not biased towards a particular class. Figures \ref{3c} and \ref{3f} show the output of Bias-Free FedGAN. We see that the outputs of Bias-Free FedGAN contains data from the minority classes (number 0 and class trousers).
We also run experiments with multiple minority classes (classes 0 and 1) and as shown in Figures \ref{4c} and \ref{4f}, we can see that our solution has images from both classes 0 and 1, while FedGAN's output (Figures \ref{4b} and \ref{4f}) are biased towards the majority class.

\section{Conclusion and Future Work}

In this paper, we show that FedGAN generates biased outcomes under non-iid settings and provide a solution, Bias-Free FedGAN, to address the same. To demonstrate the ability of Bias-Free FedGAN, we conduct experiments on the MNIST and FashionMNIST datasets. Our experiments show that Bias-Free FedGAN produces fair results under non-iid settings. Also, Bias-Free FedGAN has the same communication cost as that of FedGAN. As a part of future work, we plan to train Bias-Free FedGAN in a differentially private manner to prevent model inversion attacks and provide a detailed explanation on how privacy affects fairness and vice versa.

\bibliographystyle{plain}
\bibliography{ref}
\vspace{35em}
\appendix

\begin{algorithm}
\DontPrintSemicolon

  \KwInput{Number of training rounds $N$. Initialize local generator and discriminator for each client $i$: $\boldsymbol \alpha_0^i=\boldsymbol \alpha'$ and $\boldsymbol \beta_0^i= \boldsymbol \beta'$, $\forall i \in {1, 2, ..., M}$. Learning rates of discriminator and generator, $\eta_1$ and $\eta_2$. Noise seed, $z$. }

  \For{n = 1 to N}
  {
    Each client $i$ computes local gradient $\bm{g}^i(\bm{\alpha}^i_n,\bm{\beta}^i_n)$ from $D_i$ and $\bm{h}^i(\alpha^i_n,\beta^i_n)$ from $D_i$ and synthetic data generated by the local generator.

    Each client $i$ updates its local model in parallel to other clients via 
    \begin{equation}
        \bm{\alpha}^i_n = \bm{\alpha}^i_{n-1} + \eta_1 \bm{g}^i(\bm{\alpha}^i_{n-1},\bm{\beta}^i_{n-1})
    \end{equation}
    \begin{equation}
        \bm{\beta}^i_n = \bm{\beta}^i_{n-1} + \eta_2 \bm{h}^i(\bm{\alpha}^i_{n-1},\bm{\beta}^i_{n-1})
    \end{equation}
    
    Each client $i$ sends model parameters $\bm{\alpha}^i_n$ and $\bm{\beta}^i_n$ to aggregator
     
     Aggregator computes global generator $\bm{\alpha}^g_n$ and global discriminator $\bm{\beta}^g_n$ by averaging
     \begin{equation}
        \bm{\alpha}^g_n = \sum_{j=1}^{M} \bm{\alpha}_n^j ; \bm{\beta}^g_n = \sum_{j=1}^{M} \bm{\beta}_n^j
    \end{equation}

     Aggregator sends $\bm{\alpha}^g_n$ and $\bm{\beta}^g_n$ to clients and clients update 
     \begin{equation}
         \bm{\alpha}^i_n = \bm{\alpha}^g_n ; \bm{\beta}^i_n = \bm{\beta}^g_n , \forall i \in {1, 2, ..., M}
     \end{equation}

  }
  
\caption{FedGAN \cite{rasouli2020fedgan}}
\label{ref:fedgan-algo}

\end{algorithm}

\end{document}